\newcommand{\algoname}[1]{\textsf{{\small Concept-Monitor}}}

\newcommand{\danchor}[1]{$d_{\textrm{anchor}}$}
\newcommand{\Dprobe}[1]{$D_{\textrm{probe}}$}

\documentclass{article}

\usepackage{microtype}
\usepackage{graphicx}
\usepackage{subfigure}
\usepackage{booktabs} % for professional tables
\usepackage{url}
\usepackage{csquotes}
\usepackage{makecell}
\usepackage{float}
\usepackage{enumitem,kantlipsum}
\usepackage{wrapfig}
\usepackage{hyperref}

\usepackage[accepted]{icml2023}

\usepackage{amsmath}
\usepackage{amssymb}
\usepackage{mathtools}
\usepackage{amsthm}

\usepackage[capitalize,noabbrev]{cleveref}

\theoremstyle{plain}

\theoremstyle{definition}

\theoremstyle{remark}

\usepackage[textsize=tiny]{todonotes}

\icmltitlerunning{Concept-Monitor: Understanding DNN training through individual neurons}

\begin{document}

\twocolumn[
\icmltitle{Concept-Monitor: Understanding DNN training through individual neurons}



\icmlsetsymbol{equal}{*}

\begin{icmlauthorlist}
\icmlauthor{Mohammad Ali Khan}{equal,yyy}
\icmlauthor{Tuomas Oikarinen}{equal,yyy}
\icmlauthor{Tsui-Wei Weng}{yyy}
\end{icmlauthorlist}

\icmlaffiliation{yyy}{UC San Diego}
% \icmlaffiliation{hdsi}{UC San Diego}

\icmlcorrespondingauthor{Mohammad Ali Khan}{makhan@ucsd.edu}
\icmlcorrespondingauthor{Tuomas Oikarinen}{toikarinen@ucsd.edu}
\icmlcorrespondingauthor{Tsui-Wei Weng}{lweng@ucsd.edu}

\icmlkeywords{Machine Learning, ICML}

\vskip 0.3in
]



\printAffiliationsAndNotice{\icmlEqualContribution} % otherwise use the standard text.

\begin{abstract}
In this work, we propose a general framework called \algoname{} to help demystify the black-box DNN training processes automatically using a novel unified embedding space and concept diversity metric. \algoname{} enables human-interpretable visualization and indicators of the DNN training processes and facilitates transparency as well as deeper understanding on how DNNs develop along the during training. Inspired by these findings, we also propose a new training regularizer that incentivizes hidden neurons to learn diverse concepts, which we show to improve training performance. Finally, we apply \algoname{} to conduct several case studies on different training paradigms including adversarial training, fine-tuning and network pruning via the Lottery Ticket Hypothesis.
\end{abstract}

% Despite the successes of deep neural networks (DNNs) on a broad range of tasks little has been understood of why and how they achieve such victories due to the complexity and opaqueness of their architectures and training processes.

% For example, we find that the lottery ticket hypothesis discovers a mask that makes neurons more interpretable at initialization, \textit{without} any finetuning, and we also found that adversarially robust models rely on lower level concepts such as color than standard models trained on the same dataset. Inspired by our findings we also propose a new training regularizer that incentivizes hidden neurons to learn diverse concepts which we found to improve training performance.

% Using \algoname{}, we observe and compare different training paradigms including: standard training, adversarial training and network pruning via the Lottery Ticket Hypothesis. This brings new insights on why and how adversarial training and network pruning work and how they modify the network during training.

\section{Introduction}
Unprecedented success of deep learning has led to its rapid application to a wide range of tasks; however, deep neural networks (DNNs) are also known to be complex and non-interpretable. To deploy these DNN models in the real-world, especially for safety-critical applications such as healthcare and autonomous driving, it is imperative for us to understand what is going behind the black box.

Lots of research effort has focused on developing methods to interpret black-box DNNs, such as attributing DNN's predictions to individual input features and identify which pixels or features are the most important~\citep{zhou2016learning,gradcam,sundararajan2017axiomatic,smilkov2017smoothgrad} or investigating the functionalities (also known as \textit{concept}) of each individual-neuron (or channel of a CNN)~\citep{zeiler2014visualizing, olah2020zoom, netDissect,mu2020compositional,hernandez2022natural,clip-dissect}.

However, most of these methods only focus on examining a DNN model \textit{after} it has been trained, and therefore missing out useful information that could be available in the training process. For example, it would be very useful for deep learning researchers and engineers to understand \textit{what the concepts learned by the DNN model are} and \textit{how the concepts evolve along the training process}.

At the current stage, training DNNs is still considered a black-box and trial-and-error process. Making the training process more human-interpretable and transparent, can significantly benefit the research in deep learning because \textbf{(i)} it can shed light on why and how DNNs learn, which could be helpful to inspire new and improved DNN training algorithms; \textbf{(ii)} it can also help to debug DNNs and prevent catastrophic failure if anything goes wrong.

\begin{figure*}[t]
    \centering
    \includegraphics[width=0.95\textwidth]{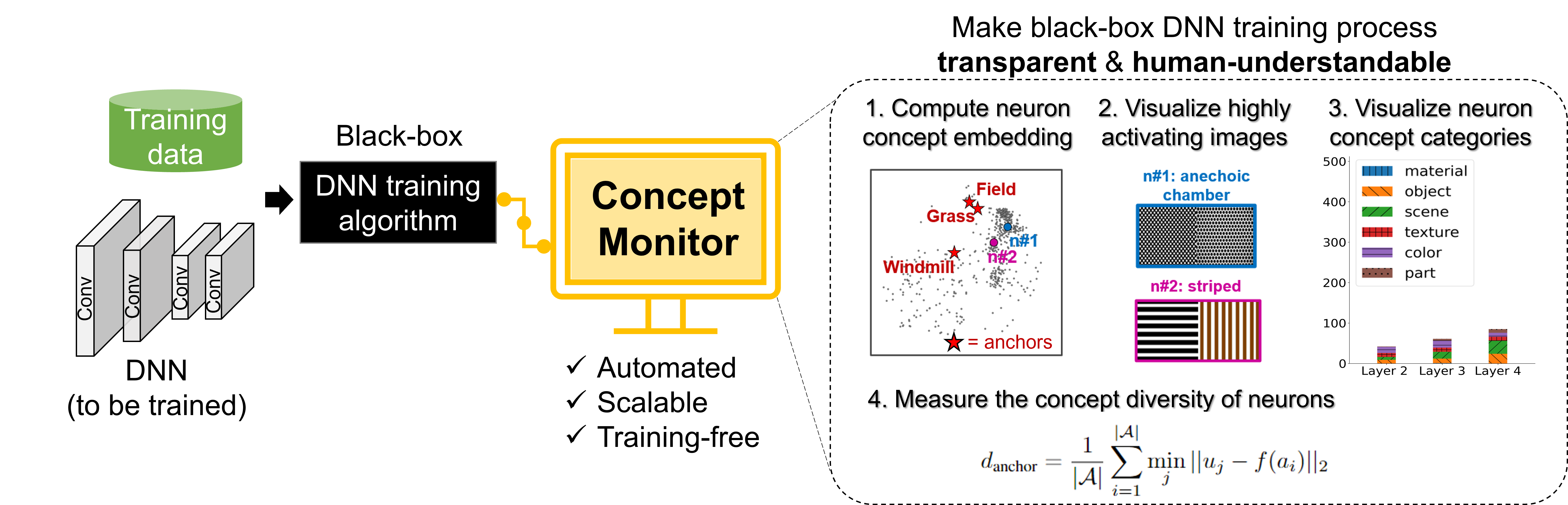}
    \caption{Our proposed \algoname{} is automated, scalable, training-free and makes DNN training process more transparent and human understandable. It consists of 4 key steps to understand each training iteration.}
    \label{fig:schematic}
\end{figure*}

Motivated by the above need, in this paper we propose \algoname{}, which is an automatic and efficient pipeline to make the black-box neural network training more transparent and interpretable. Our pipeline tracks and visualizes the training progress with human-interpretable concepts of individual neurons, which provides useful insights of the DNN model as a whole. Fig \ref{fig:schematic} gives a schematic overview of the \algoname{}. Our technical contributions can be summarized as below:
\begin{itemize}
    \item We develop a natural language based embedding space which allows us to efficiently track how the neurons' concepts evolve and visualize their semantic evolution throughout the training process. %without the need to re-learn an embedding space proposed in prior work.
    \item We provide four case studies (standard training, adversarial training, lottery ticket hypothesis and fine-tuning) to analyze various deep learning training paradigms and discover insights into how and why these alternative training processes succeed. %including training standard deep vision models, the mysterious lottery ticket hypothesis, adversarially robust training and fine-tuning on a medical dataset. With \algoname{}, we are able to
    \item We propose a quantitative metric of concept diversity to measure how diverse of a set of concepts the network is learning. Building upon this metric, we further propose a novel training modification to encourage concept diversity and show it increases accuracy and interpretability.
\end{itemize}

% \begin{figure}[h]
%     \centering
%     \includegraphics[width=0.9\columnwidth]{final_figs/Fig2_final.png}
%     \caption{Visualizing the concept evolution of Neuron 256 (blue)  and Neuron 479 in Layer 4 (purple) for standard training of a Resnet-18 model on Places365 dataset using \algoname{}.}
%     \label{fig:visConc}
% \end{figure}

\section{Background and related work}

\subsection{Neuron-level interpretability methods}

A number of methods have been proposed to understand and interpret the roles of individual neurons in neural networks, which we call  \textit{neuron}-level interpretability methods. These include visualization based methods such as \cite{erhan2009visualizing, zeiler2014visualizing, olah2017feature} and methods based on manual inspection \cite{zhou2014object, olah2020zoom, goh2021multimodal}.

More recently, methods have been proposed to automatically describe the role of neurons without need for human input, such as~\citep{netdissect2017,mu2020compositional, clip-dissect, hernandez2022natural}. Network dissection and its variation~\citep{netdissect2017,mu2020compositional} require a curated probing data labelled with pre-defined concepts. The key idea of Network Dissection is to identify \textit{concepts} of neurons by calculating an Intersection over Unit (IoU) score of intermediate activation maps and pre-defined concept masks. %as well as Test Concept Activation Vector and its variation~\citep{tcav,cace,ace2019}
 %while the key idea of Test Concept Activation Vector is to use directional derivatives to quantify the model’s sensitivity to the pre-defined concepts.

However, this approach is limited by the need of a curated probing dataset \textit{annotated} with concept labels, which may be expensive and time-consuming to collect. A new version of Network Dissection~\cite{bau2020understanding} tries to alleviate this limitation by using segmentation models to label the concepts, although the segmentation models still require dense concept labels to train. Alternatively, a recent work CLIP-Dissect \citep{clip-dissect} tries to address this challenge differently by leveraging the paradigm of multi-modal models~\citep{CLIP} to allow automatic identification of neuron concepts without the need of collecting concept labelled data or densely annotated data. We note that these techniques~\citep{netdissect2017,mu2020compositional,clip-dissect} are all compatible to our proposed \algoname{} to facilitate \textit{automatic} concept monitoring on the DNN training process. We demonstrate the versatility of our \algoname{} by showing the experimental results with different concept detectors in the Appendix for monitoring standard DNN training process. 

\subsection{Understanding DNN training dynamics} 
Most of the existing works are primarily focused on analyzing models \textit{after} training instead of investigating how the concepts change \textit{during} the DNN training process, which is the main focus of our work. A recent work~\citep{park2022conceptevo} also proposes inspecting concepts of neurons during training, which is similar to our goals, but their proposed approach is substantially different from our approach and may come with some limitations as discussed below. First, their method requires learning a universal semantic space for each neuron from a base model, while our approach does not need to perform any training. Their approach could be expensive as re-training is required when the base model or the probing dataset is changed. Second, their approach may be hard to automate, because human intervention is required to describe the behavior of each neuron, which may not be scalable to large models. In contrast, our method is fully automated and does not require human input.

 \subsection{Interpretable training}
 
 Our \textit{concept diversity regularizer} proposed in Section \ref{sec:diversitytraining} is perhaps most closely related to interpretable training methods such as Concept Whitening \cite{chen2020concept} or the learning of Concept Bottleneck Layer in \cite{oikarinenlabel}. However, these methods are technically different from our approach and focused on increasing interpretability of a NN model, while our \textit{concept diversity regularizer} is aimed to increase network performance such as accuracy. %method is mainly aimed at making DNN \textit{training process} \textit{transparent} and \textit{interpretable}, which can further help to increase network performance such as accuracy as demonstrated in our experiments.

% \section{Using \algoname{} to demystify black-box DNN training process}
\section{Methods}

In section \ref{sec:components} we detail the key components in \algoname{} including the concept detector, the unified embedding space, and the concept diversity metric. A full pipeline of how to use \algoname{} is described in section~\ref{sec:useCM}. Next in section \ref{sec:standard_train}, we use \algoname{} to demystify the standard training process of a deep vision model (shown in Fig~\ref{fig:standard_training_combined}) and discuss the results and insights.

\subsection{Components}
% \lily{merge description of network dissection (the IoU) score into this paragraph}
\label{sec:components}
\paragraph{(I) Concept Detector:} The first part of our method is to use a concept detector $\phi$ to automatically identify the concept of a neuron at any stage in the training. Given a set of concept words $\mathcal{S}$ and a probing image dataset \Dprobe, a concept detector $\phi$ would return a concept word $w^n_{1'}$ for a neuron $n$ that maximally activates it. To monitor DNN training process with automatic concept monitoring, we define a similarity metric $\texttt{sim}_i^n$ which characterize how well a neuron $n$ is described to a concept $w_i$. Note that we use the notation $w_{1'}^n$ to denote the best concept for neuron $n$ and $w_i$ to be the $i$th concept word in the concept set $\mathcal{S}$: i.e. $1' = \textrm{argmax}_i \texttt{sim}_i^n$. %Similarly, we use $w_{k'}^n$ to denote the concept word of the $k$th smallest distance $d_{k'}^n$.
This allows us to unify existing neuron-interpretability methods in this framework -- for example, the similarity $\texttt{sim}_i^n$ will be the IoU score between $n$th neurons activation map and the $i$th concept mask in Network-Dissection \citep{netDissect}, and the similarity $\texttt{sim}_i^n$ will be the similarity metric (e.g. cosine similarity, soft-WPMI) between $n$th neuron activations and the $i$th concept activations in CLIP-Dissect \citep{clip-dissect}. In addition, we say a neuron $n$ is an \textit{interpretable neuron} if $\texttt{sim}_{1'}^n > \tau$, where $\tau$ is a threshold dependent on the concept detector $\phi$, and $w_{1'}^n$ is the concept of this neuron.

% Note that since our method is general, when using a new concept detector, we only need to change the distance vector $d^n$ associated with that concept detector, which describes how closely related a neuron is to a specific concept.

% define \textit{interpretable neuron}, which are the neurons whose distance to the closest concept word is less than some threshold, i.e. $\min_i d_i^n < \tau$, where the threshold $\tau$ is dependent on the concept detector $\phi$.

% we use two representative automated neuron-level interpretability tools, Network Dissection \citep{netDissect} and CLIP-Dissect \citep{clip-dissect} as the concept detectors in our experiment as a proof-of-concept, and we note that \algoname{} is compatible with other neuron-level tools as well. Although the technical approach of each concept detector is different,
% we can actually unify them into the same framework as  calculating

\begin{figure*}[t!]
    \centering
    \includegraphics[width = 0.99\textwidth]{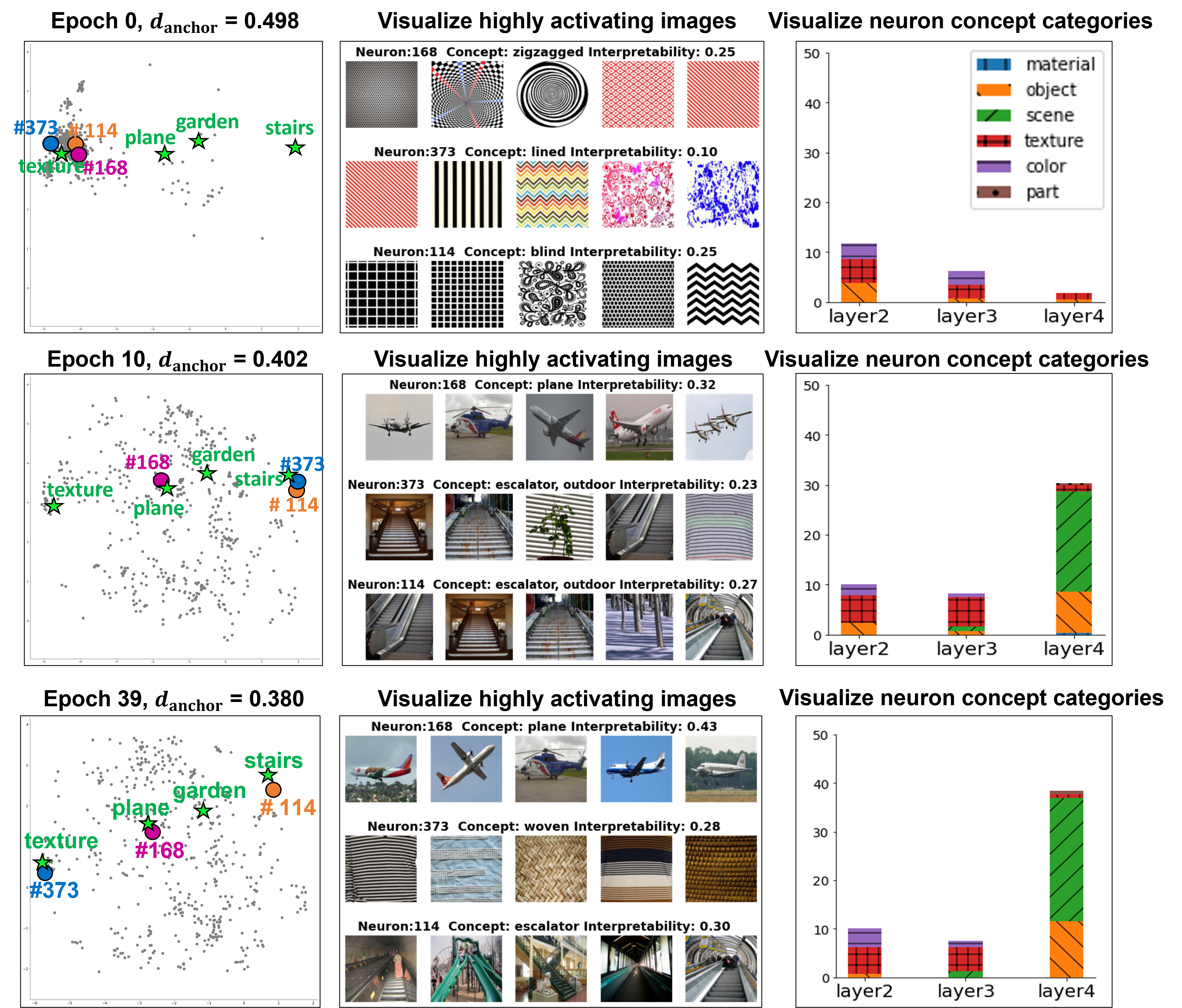}
    \caption{\textbf{Case study (I): Monitoring standard training}. We use \algoname{} to analyze a standard trained Resnet-18 model on Places365 dataset. We visualize the training at three different epochs, specifically tracking the trajectories of neurons \#114 (orange circle),\#168 (pink circle) and \#373 (blue circle) of layer 4. The 1st column plots our unified embedding space, where each gray dot represents a neuron in layer 4 and green stars represent anchor words. The tracked neurons are coloured differently for visualization. The 2nd column shows the highly activating images of the tracked neurons along with their similarity to the closest concept. Finally the 3rd column shows the percentage of interpretable neurons in layer 2-4 and which category they belong to.}
    \label{fig:standard_training_combined}
\end{figure*}

\textbf{(II) Unified embedding space:}
The second part of our method is to define a unified embedding space in order to visually track neurons' evolution. Here we detail the steps to project a neuron $n$ into our embedding space. Let $f$ be the text encoder of a pretrained large language model (e.g. the text encoder from CLIP~\cite{CLIP}). First, we compute the text embedding $v_i$ of each concept word $w_i$ in the concept set $\mathcal{S}$: $v_i = f(w_i)$. Next, we use these text embeddings $\{v_1, v_2, \dots,  v_{|\mathcal{S}|}\}$ as the basis of our semantic space and project neurons into this space using a weighted linear combination of $v_i$. We use the concept detector $\phi$ to compute $\texttt{sim}_i^n$ for all concept words $w_i$ and neurons $n$, which are subsequently used to calculate weighting using softmax with temperature $T$. The embedding $u_n$ of neuron $n$ is then calculated as:
\begin{equation}
    \label{eq:embedding}
    u_n = \sum\limits_{i=1}^{|\mathcal{S}|} \lambda_i^n f(w_{i}), \;\; \lambda_i^n = \frac{e^{\texttt{sim}_{i}^n/T}}{\sum_{j=1}^{|\mathcal{S}|} e^{\texttt{sim}_{j}^n/T}}
\end{equation}
where $\lambda_i^n$ is the weight describing the similarity of concept $w_i$ to neuron $n$. Finally, we visualize the concept embedding $u_n$ in two dimensional space using UMAP \cite{mcinnes2018umap-software} in plot such as Fig \ref{fig:standard_training_combined} (column 1). We can use the concept embedding plot in the unified embedding space to track the concept evolution of each neuron easily.

Another benefit of our unified embedding space is that we can project any general concept word $\alpha$ into the same embedding space by calculating its text embedding $f(\alpha)$. This lets us mark the embedding space with concept "anchors" (see the green stars in Fig \ref{fig:standard_training_combined}, column 1), which could be concepts that a researcher thinks should exist in a well trained model, or undesirable concepts such as ones representing bias. With the concept-anchors, researchers can then track whether and which neurons are converging or diverging away from anchors.%, which could give useful feedback during training -- we will describe more in Section~\ref{sec:diversitytraining} on how to design new training methods based on \algoname{}.

% For example, if the researchers don't want the model to represent some spurious concept in the model, like presence of grass in a horse detector, they can then use the "grass" and related anchors to check whether some neuron's are converging towards it.

\textbf{(III) Concept diversity metric}

Another benefit of our neuron concept visualization via unified embedding space (e.g. Fig \ref{fig:standard_training_combined}) is that it allows us to easily sense the diversity of concepts represented by the neurons. As we will see in Sections \ref{sec:standard_train}, neurons of well trained models typically cover a large set of concepts, while poor training typically leads to a lack of concept diversity. Inspired by this, we propose a quantitative metric \textit{anchor distance} to measure concept diversity based on our unified embedding space. The idea behind this metric is that we have a set of text anchors $\mathcal{A} = \{a_1, a_2, ..., a_{|\mathcal{A}|}\}$ that ideally describes concepts that are important for this task, and we want to make sure at least one neuron in the network has a concept similar to each anchor. Thus, we define the \textit{anchor distance} $d_{\textrm{anchor}}$ as follows:
\begin{equation}
    \label{eq:anchor_loss}
    d_{\textrm{anchor}} = \frac{1}{|\mathcal{A}|} \sum_{i=1}^{|\mathcal{A}|} \min_{j} ||u_j - f(a_i)||_2
\end{equation}
where $u_j$ are the neuron embeddings as defined in Eq. \eqref{eq:embedding}. This metric measures the average Euclidian distance of the closest neuron to each concept in the anchor set. As such, networks with highly diverse neurons will reach low \textit{anchor distance} while highly clustered neurons will results in a large $d_{\textrm{anchor}}$. For most of our experiments we set $\mathcal{A} = \mathcal{S}$,  where both of them are the set of labels in Broden dataset, but they can be easily changed based on task and user needs.

\subsection{Using Concept-Monitor}
\label{sec:useCM}
With all the components in place, we now describe how they come together to form \algoname{} for DNN training. \algoname{} offers a combination of metrics and visualizations to analyze a snapshot of a model, and by analyzing at consecutive snapshots we can understand how the model evolves in terms of concepts learned by individual neurons. The schematic of \algoname{} is illustrated in Figure~\ref{fig:schematic}. At each snapshot, concept monitor produces the following:
\begin{enumerate}
    \item Two dimensional plot of neuron embeddings, and select text anchors (marked as the stars in the plot).
    \item Visualization of detected concepts and most highly activating images for each neuron.
    \item Bar plot visualization of the number of interpretable neurons and which category they belong to.
    \item A numeric measure of concept diversity $d_{\textrm{anchor}}$ as defined in Eq. \eqref{eq:anchor_loss}.
\end{enumerate}

A visualization of all these information for different epochs of standard models is shown in Figure \ref{fig:standard_training_combined}. We think this information should be combined with standard metrics such as accuracy and losses to monitor training progress in an interface similar to TensorBoard \cite{tensorflow2015-whitepaper}. This information can be helpful in several ways, for example, if we see that concept diversity starts to decrease, it indicates issues in the training. In Sections \ref{sec:standard_train} and \ref{sec:case_studies} we apply concept monitor on different training runs, and in section \ref{sec:diversitytraining} we propose a training modification inspired by our insights from \algoname{}.

\subsection{Case study (I): Monitoring standard training}
\label{sec:standard_train}

Now we use \algoname{} to investigate standard training of ResNet-18 model on Places365 dataset with CLIP-Dissect as the concept detector. We study how the concepts evolve across training and whether there is a correlation between accuracy and concept generalization with \algoname{}. We investigate the concept evolution of neurons at different epochs in the training process using the full pipeline described in Section~\ref{sec:useCM}. The main results are plotted in Fig \ref{fig:standard_training_combined}, where row 1 represents a very early snapshot (trained for 1 epoch, at the end of Epoch 0), row 2 an intermediate snapshot (Epoch 10) and row 3 the final snapshot (Epoch 39, training ends). The 1st column visualizes our proposed unified embedding: we use green star symbols to show the anchors embeddings $f(a_i)$ for anchors [texture, plane, garden, stairs], and we use the circle symbols to show 3 neuron embeddings $u_n$ -- neuron \#114 (colored orange), \#168 (colored pink) and \#373 (colored blue). The 2nd column visualizes the highly activating images for these 3 neurons, which should match the detected concept (word) displayed on top of it (e.g. neuron \#168 in Epoch 0 has detected concept "zigzagged" on the graph). The 3rd column visualizes the percentage of interpretable neurons from the 6 pre-defined categories. Now we summarize three observations from the standard training below:
\setlength{\itemsep}{0pt}
\begin{enumerate}[leftmargin=*]
    \item \textit{Model learns to look at more complex features as training progresses.} - As shown in Figure \ref{fig:standard_training_combined} second column, initially all neurons start by detecting low level features like lines, patterns and textures. This can also be seen as all neurons being clustered around the \emph{textures} anchor in the embedding space (column 1) and absence of 'object' and 'scene' categories (column 3). We see that as training progresses, the model starts to learn more complex features which can be clearly seen from the highly activating images in column 2 and bar plots in column 3.

    \item \textit{Shallower layers learn more low-level features like material and texture while deeper layers learn more nuanced object detectors.} - We consider the broad categories of [\textit{Material, Object, Scene, Texture, Color, Part}] to group neurons similar to the labels used in Broden dataset. We note  that the categories \textit{Scene}, \textit{Object} and \textit{Part} are concerned with higher level concepts like \textit{planes} and \textit{stairs} while \textit{Texture} and \textit{Color} are concerned with lower-level concepts like \textit{lined}, \textit{zigzagged} etc. From column 3 in Figure \ref{fig:standard_training_combined}, it's evident that Layer 2 and Layer 3 are learning a lot more low level information than Layer 4 as the texture and color neurons are represented more.

    \item \textit{Concept diversity happens relatively early in the training.} - Using the unified embedding space, we can see that the neurons are clumped together initially in the embedding plot and as training progresses they spread out eventually converging to their finalt concept. However, we note that this divergence occurs during the earlier stages of the training and after that the neurons mostly stay close to their concepts. For example in Figure \ref{fig:standard_training_combined} column 1, we see that neuron \#114 (orange circle) and neuron \#168 (pink circle) reach and retain their position for the last 30 epochs of training and keep their concepts as seen in highly activating images.
 \end{enumerate}

 \paragraph{Discussion of standard training.} Using our method in standard training, we have seen a correlation between training stage and interpretability of a model (represented as the number of interpretable neurons in the Column 3 of Fig \ref{fig:standard_training_combined}).
 We notice that for a well trained model, there is a progression from a low level concepts understanding to higher level conceptual understanding, and the concept diversity increases as training progresses. We also run \algoname~ using Network Dissection \citep{netDissect} and MILAN \citep{milan} as the concept detectors in Appendix (see Fig \ref{fig:netdissect} and Fig \ref{fig:milan}) and show our observations are consistent across different concept detectors.

\begin{figure}[h!]
     \centering
     \includegraphics[width=1.0\linewidth]{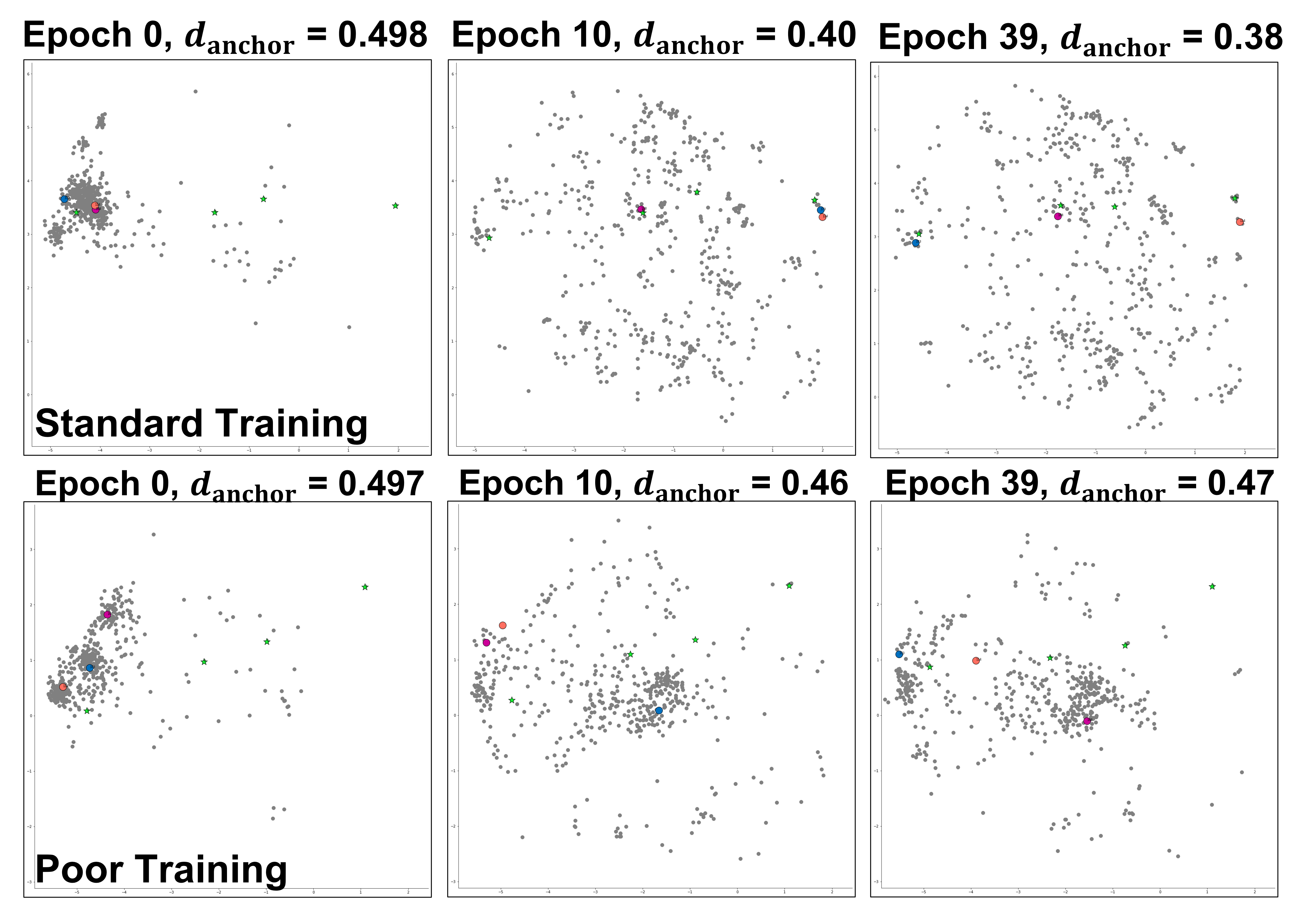}
     \caption{Investigating poor training using unified embedding space. We see that in poor training the neurons don't diverge much along training. The high \danchor{} also indicates inability to represent useful concepts.}
     \label{fig:poor-training}
 \end{figure}

\paragraph{Poor training.}
% \label{sec:poortraining}
    Next, we use \algoname{} to investigate standard training with poor hyper-parameter selection, specifically the standard training in Section 3.3 but with a fixed high learning rate. As expected, the large learning rate made it impossible for the model to train properly with the final accuracy being $~3\%$. We visualize the unified embedding plot of the poorly trained model snapshots (in the 2nd row) and contrast it with the standard trained model snapshots (in the 1st row) in Figure \ref{fig:poor-training}. It can be seen that unlike standard training, the concept diversity doesn't increase in the poor training but instead plateaus, which means the poorly trained model has a higher \danchor. This is also confirmed with the calculated \danchor~ value: \danchor~ for poor training is $0.47$ compared to $0.38$ in the well-trained standard model, showing the inability of the model to learn diverse concepts.

\vspace{-3mm}
\section{Concept Diversity Regularizer}
\label{sec:diversitytraining}

In Section \ref{sec:standard_train}, we find that the neurons of well trained models usually cover a large variety of concepts while poor training often leads to neurons clustering together. Inspired by this, we propose a regularizer to increase concept diversity based on the \textit{anchor distance} \danchor{} from Section \ref{sec:components}. We find that it can improve model accuracy, interpretability and concept diversity as shown in Section \ref{subsec:trainresult}.

In order to include the \textit{anchor distance} in training, \danchor{} needs to be differentiable. By plugging the neuron embeddings $u_j$ in Eq. \eqref{eq:embedding} into Eq. \eqref{eq:anchor_loss}, we can rewrite \danchor{} as below:
\begin{equation}
    \label{eq:anchor_loss_rewrite}
    \frac{1}{|\mathcal{A}|} \sum_{i=1}^{|\mathcal{A}|} \min_{j} ||\sum\limits_{i=1}^{|\mathcal{S}|} (\frac{e^{\texttt{sim}_{i}^j/T}}{\sum_{k=1}^{|\mathcal{S}|} e^{\texttt{sim}_{k}^j/T}}) \cdot f(w_{i}) - f(a_i)||_2.
\end{equation}
Recall that $f$ is a text encoder of a pretrained large language model, and $\texttt{sim}_i^j$ is a similarity metric that characterize how close a neuron $j$ is related to a concept $w_i$ based on the selected concept detector $\phi$. In our setting, $f$ is frozen and fixed while $\texttt{sim}_i^j$ contains the trainable parameters of the DNN  classifier model (through $\phi$). Thus, if $\texttt{sim}_i^j$ can be made differentiable, then \danchor{} becomes a differentiable function of the DNN parameters, and as such can be used be used as a regularizer. In Network Dissection, $\texttt{sim}_i^j$ is an IoU score that is not differentiable, but in CLIP-Dissect, $\texttt{sim}_i^j$ can be made differentiable if using the differentiable \textit{cos cubed} similarity function defined in \cite{oikarinenlabel}. Thus, we use the \textit{cos cubed} similarity function and introduce the concept diversity regularizer directly in the training to reduce the \danchor:
 \begin{equation}
    \label{eq:joint_loss}
     L = L_{\textrm{std}} + \beta d_{\textrm{anchor}}
 \end{equation}
where $L_{\textrm{std}}$ is the standard loss such as cross-entropy, \danchor{} is defined in Eqn. \eqref{eq:anchor_loss_rewrite} and $\beta$ is a hyper parameter to decide the relative importance of the two losses. Our goal is to minimize Eq.~\eqref{eq:joint_loss} and learn a model that has a reduced \textit{anchor distance}.

% Specifically, \danchor{} is differentiable if the neuron embeddings $u_j$ are differentiable.

% In section \ref{sec:components} we proposed \textit{anchor distance} \danchor{} as a way to measure concept diversity.

% The key insight of this section is that in some cases \textit{anchor distance} is a differentiable function of the DNN parameters, and as such can be used be used as a regularizer.

\subsection{Results}
\label{subsec:trainresult}

To test the performance of our regularizer, we train a model on a subset of Places365 like we did in section \ref{sec:standard_train}. We trained it exactly like before, using a mini-batch stochastic gradient descent as the optimizer but with the new joint loss function of Eq. \eqref{eq:joint_loss}. The regularizer \danchor{} was calculated over the neurons in the second to last layer(layer4) of the network. Typically we calculate \danchor{} using the Broden dataset, but this would be too expensive to compute during training. Instead we simply use a minibatch of Places training data as \Dprobe{} which greatly reduces computational overhead at the cost of introducing more noise to the neuron embeddings. We used $\beta=1$ for our experiments.

\begin{table}[h!]
\centering
\scalebox{0.85}{
\begin{tabular}{@{}lcccc@{}}
\toprule
Model & Accuracy & \begin{tabular}[c]{@{}l@{}}\# Neurons\\ interpretable\end{tabular} & $d_{\textrm{anchor}}$ \\ \midrule 
Res18 std & $47.48 \pm 0.13\%$ & $106.33 \pm 4.04$ & $0.39 \pm 0.007$ \\ 
Res18 ours & $\mathbf{48.33 \pm 0.08\%}$ & $\mathbf{147.7 \pm 7.02}$& $\mathbf{0.36 \pm 0.004}$ \\ \midrule \midrule
Res50 std & $49.12 \pm 0.44\%$ & $\mathbf{544.4 \pm 41.02}$ & $0.31 \pm 0.003$ \\ 
Res50 ours & $\mathbf{49.32 \pm 0.02\%}$ & $361 \pm 15.7$ & $\mathbf{0.27 \pm 0.006}$ \\
\bottomrule
\end{tabular}
}
\caption{Comparison between standard models and models trained with our Concept Diversity Regularizer on Places365.}
\label{tab:regularizer_results}
\end{table}
% \begin{tabular}[c]{@{}l@{}}Concept \\ Regularizer\end{tabular}

Table \ref{tab:regularizer_results} shows that training with our regularizer improved average test accuracy by ~1\% point over the standard well-trained Resnet18 model with no further optimization beyond adding the regularizer. In addition, it increased the concept diversity and number of interpretable neurons on the second to last layer. Note that the increase in interpretability is not caused by overfitting as we used Broden as \Dprobe{} in Table \ref{tab:regularizer_results}, which was not used during training. We also see a small improvement in average accuracy for Resnet50 in addition to increased concept diversity. The results of this initial experiment show the promise of encouraging concept diversity in training.

\section{Case studies of other training paradigms}
\label{sec:case_studies}
\begin{figure}[t]
    \centering
    \includegraphics[width=0.7\columnwidth]{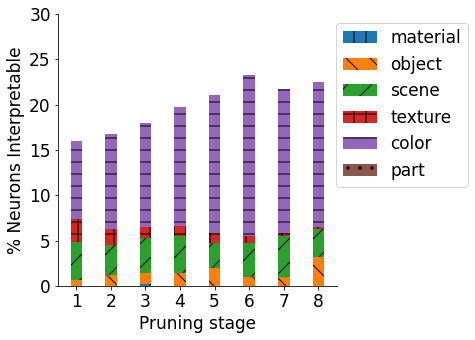}
    \caption{A barplot of the percentage of neurons that are interpretable after successive stages of pruning weights and rewinding remaining weights back to their initializations in our LTH experiment. We can see the number of interpretable neurons increases after simply setting some weights to 0. }
    \label{fig:lth_barplot}
\end{figure}

\begin{figure*}[t]
    \centering
    \includegraphics[width=0.7\textwidth]{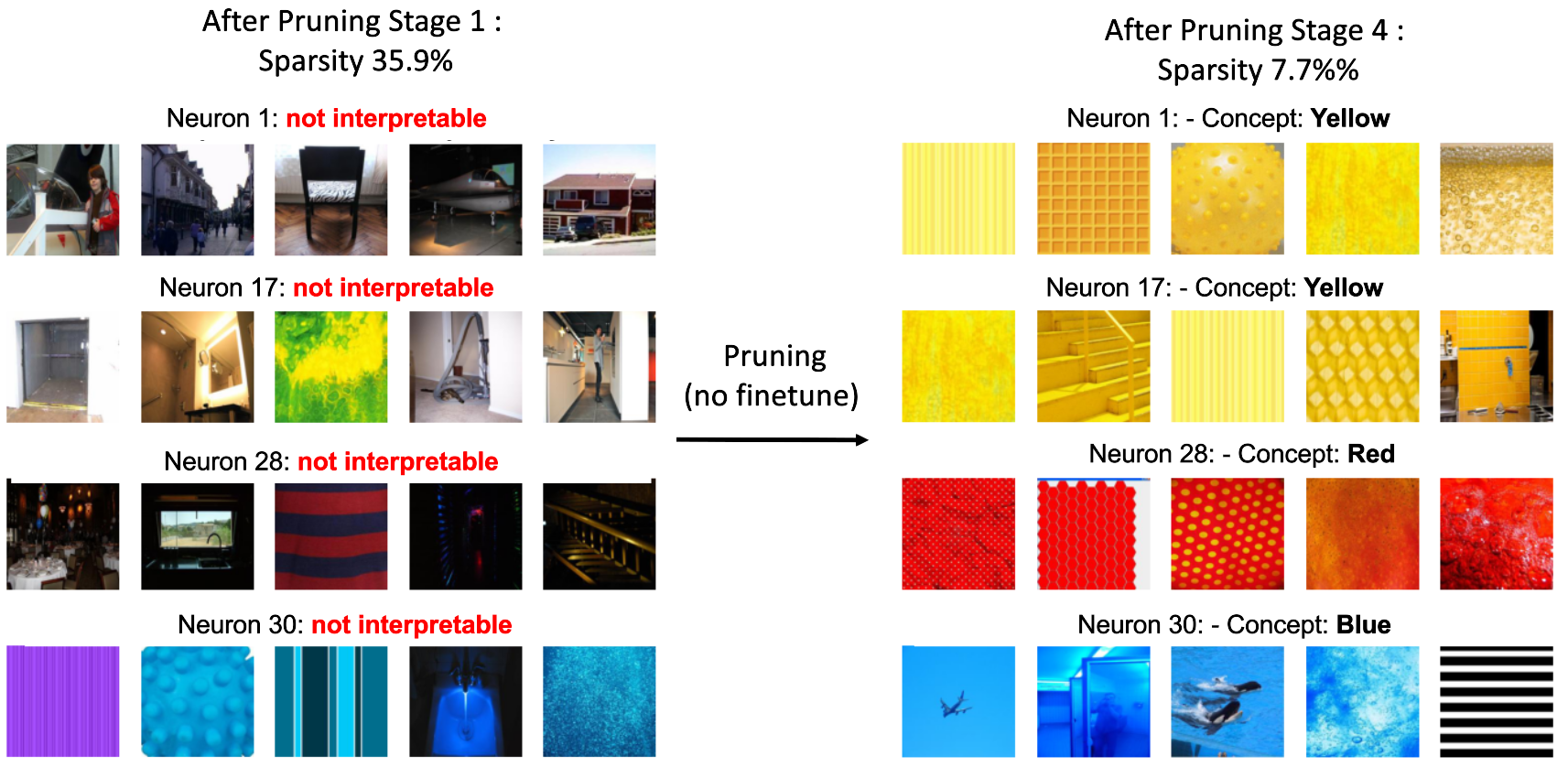}
    \caption{Select neurons from layer4 of our LTH rewind to zero model that were uninterpretable at after one stage of pruning (and rewind to initialization) but became interpretable by the end of 4th pruning stage. Note no weights were changed between the two, interpreatbility was caused purely by zeroing some weights.}
    \label{fig:lth_images}
\end{figure*}

In this section, we show that \algoname{} is versatile and can be used to study various training paradigms to gain insights into how and why they work. We also provide useful observations and insights that could help future researchers better understand these training procedures.
\subsection{Case study (II): Lottery ticket hypothesis}

Lottery Ticket Hypothesis (LTH) \citep{LTHoriginal} is a popular method to prune DNNs without sacrificing their performance. In this case study, we use \algoname{} to better understand the success behind LTH. The main idea of LTH is to use iterative magnitude pruning (IMP) to prune the model by repeating the steps of training, pruning and rewinding to an initial epoch. LTH hypothesizes the existence of "winning tickets" at initialization which are sub-networks within the network that can be trained to performance equivalent to the original model. It has been noted in literature that the process of pruning encodes some information in the pruning mask \cite{zhou2019deconstructing}. To understand this, we use \algoname{} to investigate LTH through the lens of interpretability. We train a Resnet18 on CIFAR 10 dataset using IMP in 8 stages. For full details on our experimental setup, please refer to Appendix A. A related study was done by \citep{lthinterp} which found that pruning doesn't affect the interpretability of the model until there is a significant drop in accuracy. For our experiment we focus on the original Lottery Ticket Hypothesis, where after each stage of pruning we rewind the weights all the way back to their initialization.
 %We study LTH with three different rewinding stages of IMP: rewinding to initial weights (rewind\_0), epoch 5 (rewind\_5) and epoch 16 (rewind\_16) respectively. We consider multiple runs to verify the validity of our observations but we show one results from a single run here. We use \algoname{} to track the training process of these 3 different rewinding strategies and plot the results in Fig \ref{fig:lth_barplot}.

\vspace{-3mm}
\paragraph{Observations and Discussion.}

In our analysis, we make the following observation:
\textit{Pruning the network learns to encode some concepts without any fine tuning.}
    Fig \ref{fig:lth_barplot} shows the fraction of interpretable neurons in layer 4 of the model \emph{after} each stage of pruning and rewinding. We notice that even though we are rewinding to the initial weights, the number of interpretable neurons \emph{increases}. Since the weights are randomly initialized, the only way there can be a gain in interpretable neurons is through the changes that happen during pruning, i.e. the zeroing out of certain weights in the network.
    We believe that the model may be learning to remove connections that are harming the network, which leads to neurons "learning" simple interpretable concepts such as colors already at initialization.
    Interestingly, we also notice the number of texture neurons decreases as pruning progresses. We note that a related phenomenon was observed by \citep{zhou2019deconstructing} who find that IMP zeros out weights that would ultimately go towards zero anyway after training. Hence, they hypothesize that a pruned initial network encodes a portion of the training process itself, which they refer to as "masking is learning". This could explain why we see interpretable neurons with just pruned initial weights. Further proof of this can be illustrated in the experiment result in Figure \ref{fig:lth_images}, which shows some initially uninterpretable neurons (left panel) that 'learn' to encode simple concepts such as colors through pruning only and become interpretable (right panel).

\vspace{-3mm}

\subsection{Case study (III): Adversarial Training}
\begin{figure}[t]
    \centering
   \includegraphics[width=0.99\columnwidth]{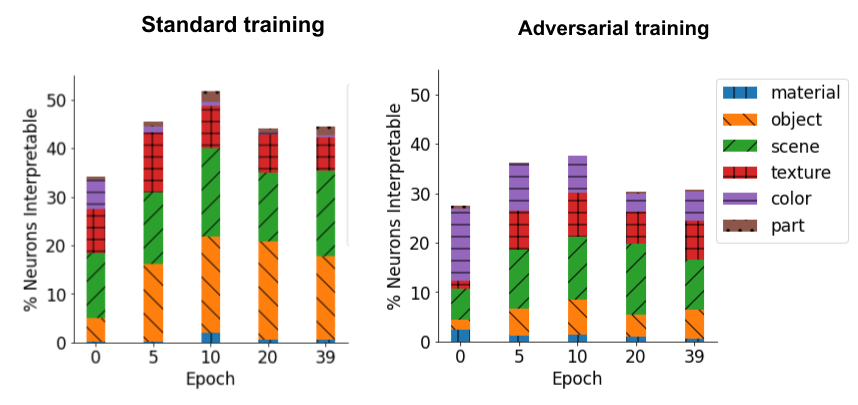}
    \caption{Comparison of the types of concepts learned by standard training compared to adversarial training in the second to last layer (layer4), and differences in types of concepts learned by the models. Note these figures are the network at the end of each epoch, so epoch 0 is after one epoch of training.}
    \label{fig:adv_bar_plot}
\end{figure}

\begin{figure*}[t]
    \centering
    \scalebox{1}[0.7]{\includegraphics[width=0.72\textwidth]{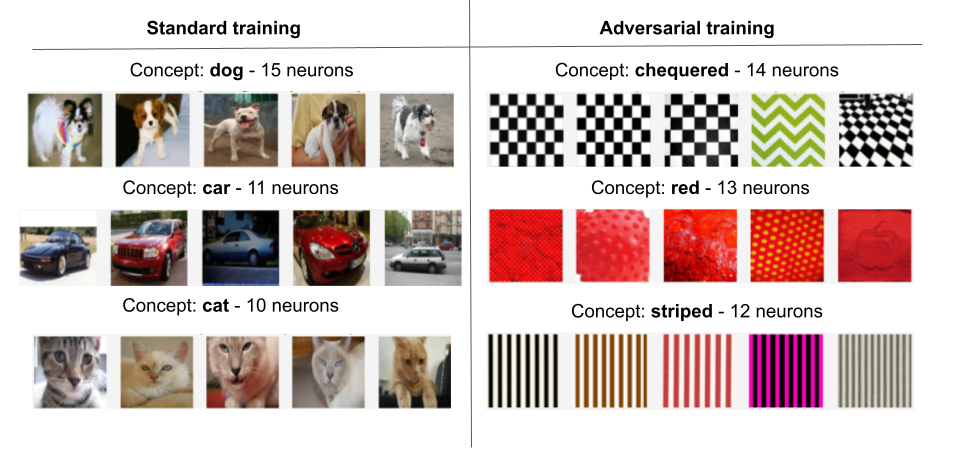}}
      \setlength{\belowcaptionskip}{-10pt}
    \caption{Examples of the most common concepts detected by layer4 of Resnet-18 for both standard and adversarial training. We can see a large difference between the types of concepts encoded, where standard training learns many neurons already detecting CIFAR-10 classes such as car, while adversarial model detects simple patterns and colors.}
    \label{fig:adv_train_examples}
\end{figure*}

DNNs are known to be vulnerable against small perturbations in their inputs \citep{szegedy2013intriguing}. This is problematic as networks can fail unexpectedly after small random or adversarial perturbations which raises concerns over their safety. Fortunately, methods have been developed to defend against adversarial attacks, most popular of these being Adversarial Training \citep{madry2018towards}. This successfully makes networks more robust against such attacks, but comes at a cost of degraded performance on clean test data. In this study, we apply \algoname{} to adversarial training to better understand how adversarial training changes a network and why standard accuracy suffers. We analyse a Resnet18 model trained on CIFAR10 \textit{with} and \textit{without} adversarial training. For full details on our experimental setup, please refer to Appendix section A.

\textbf{Observations and Results}: Using \algoname{} we have the following two observations from Figure \ref{fig:adv_bar_plot} and \ref{fig:adv_train_examples}. These observations are consistent across multiple trained models but for simplicity we focus our discussion on one model and its visualization.
\vspace{-3mm}
% \lily{should mention our results are consistent across multiple runs}
\begin{enumerate}[leftmargin=*]
      \setlength{\itemsep}{0pt}
  \setlength{\parskip}{0pt}
    % \item  \textbf{Adversarially robust network has less interpretable neurons in late layers, but more in earlier layers.}
    % In Fig \ref{fig:adv_bar_plot}, we plot the number of interpretable neurons in layer 3-4 at three different training stages. It can be seen that at the end of training that 293 out of 512 of the layer 4 neurons are interpretable for standard training (left panel) while only  215 out of 512 are interpretable for the robustly trained model (right panel). For layer 3, it is 91 out of 256 for standard model and 125 out of 256 for the robust model. We observe
    % layer 2 neurons have similar trend to layer 3, please refer to Fig \ref{fig:adv_stats_large} in Appendix for full statistics.
    \item \textit{Adversarially robust network keeps relying on colors, standard model moves on to higher level concepts.}
    In Figure \ref{fig:adv_bar_plot} we observe that robust model has a lot more interpretable neurons dedicated to detecting "color" (the purple bar) than the standard model (30 vs 2) at the end of training. On the other hand, the trained robust model has less interpretable neurons in the object scence and part categories. This finding is sensible these categories more often rely on high frequency patterns that are easily affected by $l_{\infty}$ noise, therefore the adversarial training forces the model to rely less on them and rely more on more resilient features like color. Interestingly, early in their training the concepts of the two models look more similar but they start diverging after epoch 5.
    \item \textit{Standard training develops many neurons detecting target classes in the second to last layer, robust training does not.}
    As seen in Fig \ref{fig:adv_train_examples}, the standard network has many neurons detecting the target classes of CIFAR-10 present in the second to last layer. For example, the fully trained standard network has 15 interpretable neurons detecting dogs, 11 neurons for car and 10 neurons detecting cats in layer4. In comparison the robust network has 1,3,3 respectively which indicates a limited capacity to represent these classes. The standard model learns to rely on target class neurons early on in training, with many of them present by epoch 10.
\end{enumerate}

\vspace{-3mm}
\textbf{Discussion}:
We find that adversarial training harms the ability of the network to detect higher level concepts. Since these concepts are necessary for many tasks, losing them may be a significant cause for the degradation in standard performance. This may also be related to the robust networks inability to detect target class objects in second to last layer.

On the other hand, the features learned by the robust network are more general and less task specific, as seen by larger diversity in concept types in Fig. \ref{fig:adv_bar_plot} and lack of target classes in Fig. \ref{fig:adv_train_examples}. This could explain why \citep{salman2020adversarially} found adversarially robust models to have better features for transfer learning. In effect standard model features could be overfitting to the training task.

Finally we provide the full details of \textbf{Case study (IV)} of using \algoname{} to monitor fine-tuning of a pretrained DNN in Appendix.

\vspace{-3mm}

\section{Conclusions}

We have presented \algoname{}, a novel method to automatically track and monitor neural network training process in a transparent and human-understandable way. We have demonstrated how to use \algoname{} to monitor and further improve standard training with a novel concept diversity regularizer. Additionally, with the 4 comprehensive case studies on various deep learning training paradigms, we show that \algoname{} allows us to demystify and better understand the underlying mechanism of DNN training.

% . Lottery Ticket Hypothesis and adversarial training, as well as model fine-tuning on medical task. With \algoname{}, we discover that surprisingly lottery ticket hypothesis prunes the network in a way that the neurons are interpretable even at initialization, discovering interpretability hidden in random initialization. Furthermore, we discover that adversarial training causes the hidden neurons to detect more simple concepts like colors while losing representations of materials and target class objects. We also test our method on medical dataset and find that the model learns to focus more on low level features which reflect the medical dataset.\\ 

%\textbf{Reproducibility statement:} We acknowledge the importance of replicating our experiments and for that reason we have explicitly mentioned the implementation details of all our experiments in Appendix A. 

\bibliographystyle{icml2023}

\clearpage
\appendix
\pagebreak
\appendix
% \section*{Appendix}

% \subsection{Comparison Table between \cite{park2022conceptevo} and \algoname{}}
% \input{table/concept-evo-comparison}

\section{Experimental setup}
\subsection{Case study (I): Standard training}

\underline{Setup}: We train a Resnet-18 model on Places-365 dataset, which contains a lot of diverse classes allowing the DNN model to learn diverse concepts. To reduce the training time, we randomly selected 1000 images for each of the 365 classes and trained for 40 epochs reaching top-1 accuracy of 47.5\%. We use batch size of 256 and an initial learning rate of 0.1 with cosine annealing scheduler.

\underline{Probing methodology}: We use Broden \citep{netDissect} dataset as $\mathcal{D}_{probe}$ and use associated concept labels as a decoupled concept set $\mathcal{S}$. Our embedding space, as described in section 3.1, is computed using CLIP's text embeddings of Broden labels as a basis. We used CLIP-Dissect with cosine-cubed similarity function and neuron embedding temperature $T=0.01$.
%For visualizing in a 2-dimension plot, we follow \citep{park2022conceptevo} and use UMAP dimensionality reduction \citep{mcinnes2018umap-software}, as it preserves inter-point distance in the lower dimensions.

\subsection{Case study (II): Lottery ticket hypothesis experiments}

% \begin{figure}[h]
%   \centering
%   \includegraphics[width=0.95\columnwidth]{final_figs/lth-just0.png}
%   \caption{Analysing Layer 4 for Resnet-18 using IMP at different stages of pruning. We observed IMP with rewinding to initial weights. The left plot is the accuracy vs training stages, The red dots represent accuracy after rewinding, the green dots represent accuracy after finetuning and pruning. The right plot is the number of interpretable neurons found at different training stages.}
%   \label{fig:pruning_fig}
% \end{figure}

\underline{Setup}: We train ResNet 18 on CIFAR 10 dataset using IMP as in the LTH paper \citep{LTHoriginal}, rewinding to different initial weights. For each stage of IMP we train the model for 160 epochs, prune 40\% of the weights and rewind to initialization. After 8 stages we got an accuracy of $91.18\%$ on the validation set as compared to $94.32\%$ after stage 0. Please refer to \citep{chen2022coarsening} implementation for reference.

\underline{Probing methodology}: For our $\mathcal{D}_{probe}$, we use Broden as the probing dataset and for concept set $\mathcal{S}$ we use broden labels. We use CLIP-Dissect with SoftWPMI similarity function and embedding temperature $T=0.1$.

\subsection{Case study (III): Adversarial Learning experiments}

\begin{figure}[ht]
  \centering
     \includegraphics[width=0.8\columnwidth]{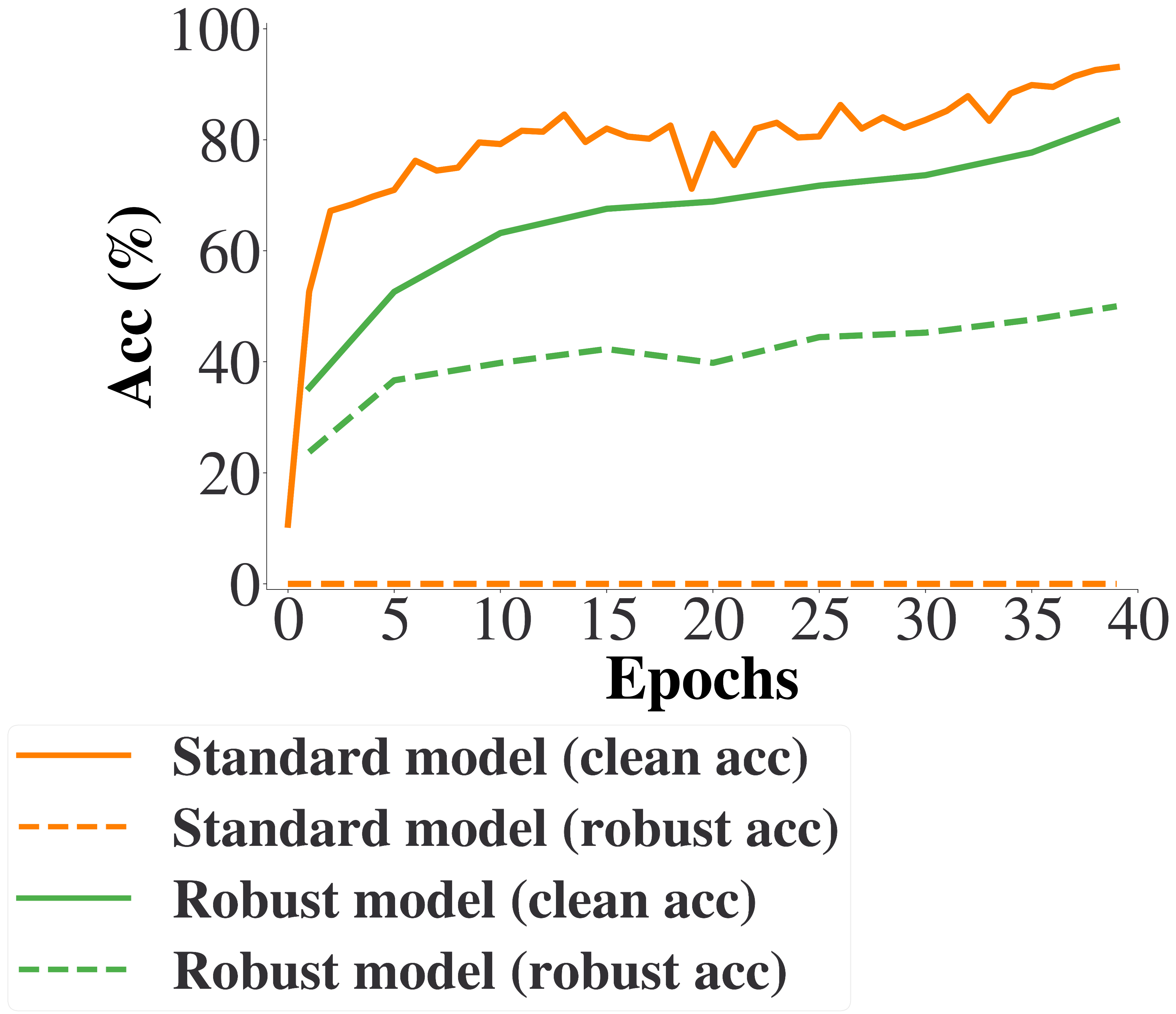}
    \caption{Accuracy vs Epoch for standard and robust model }
    \label{fig:adv_acc}
\end{figure}
\begin{figure}[ht]
\centering
   \includegraphics[width=0.8\columnwidth]{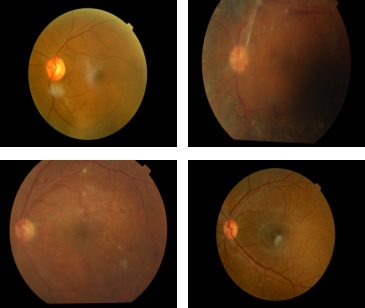}  \caption{Sample images in the diabetic retinopathy detection training dataset. The key features (e.g. dots and texture) are being detected by the interpretable neurons in Fig~\ref{fig:medical}.}
    \label{fig:trainimg}
\end{figure}

\underline{Setup}: We perform adversarial training with PGD attacks on a ResNet-18 architecture. We follow reop \citep{wong2020fast} and train the network with $\epsilon=8/255$ and $l_{\infty}$ perturbations for 40 epochs. We compare it against a CIFAR-10 network trained using the same exact training setup but no adversarial training. The standard model reaches a final accuracy 94.29\%, while the robust model reaches 83.42\% accuracy on clean data and 50.00\% robust accuracy against a PGD adversarial attack as shown in Fig \ref{fig:adv_acc}. The standard model expectedly has accuracy close to 0\% on adversarial images.

\underline{Probing methodology}: We use the same probing methodology as in Case study (II), Broden images as $\mathcal{D}_{probe}$ and for concept set $\mathcal{S}$ we use the broden labels as the concepts can be easily categorized. We use CLIP-Dissect with SoftWPMI similarity function and $T=0.1$. \\

\subsection{Case study (IV): Fine-tuning on medical dataset}

\underline{Setup}: We used ResNet-34 backbone pretrained on ImageNet dataset as our feature extractor and used a simple linear layer as the classification head. We trained this network on the diabetic retinopathy classification dataset \citep{eyedata} (Fig \ref{fig:trainimg}) and it achieved an accuracy of 72.77\%. We followed the work from \citep{diabeticRetinopathy} for our experiments. We use Broden as $\mathcal{D}_{probe}$ and broden labels as $\mathcal{S}.$

\section{Results of Case study (IV) fine-tuning on a medical dataset}
In this section, we use \algoname{} to observe the fine-tuning of a pretrained DNN on a diabetric retinopathy dataset \citep{eyedata}. The purpose of this experiment is to show that \algoname{} can be applied to a different domain such as medical data and gather insights into the process of finetuning a pretrained model. The setup details are in Appendix A.

\textbf{Observations and results}:
We observe that for the initial weights, as the neurons are pretrained on Imagenet, they show a lot of diverse and high level concepts (as shown in highly activating image of epoch 0 in the Fig \ref{fig:medical} in Appendix).  However as the training progresses, we notice that more neurons get activated by textural concepts like dots and patterns rather than objects. This is what we expect because as the model gets better at classifying retinopathy images shown in Fig \ref{fig:trainimg} (in the Appendix), we expect it to rely more on textures and presence of "dots" which is consistent to what we observe here as shown by the highly activating images of the top interpretable neurons in epochs 5 and 29 in Fig \ref{fig:medical}. From Fig \ref{fig:medical} we can also see that the number of interpretable neurons in the higher level categories like "object" and "scene" category decrease and the interpretable neurons in the "texture" and "material" category remain the same or increase. This further confirms our theory that the model learns to focus on the textural aspect of images more. This is also confirmed by the semantic embedding space where we see the neurons becoming less separated. We further confirmed this by calculating the average pairwise distance between neurons which decreased from $0.735$ to $0.726$.

% \begin{figure}[h]
% \centering
%     \includegraphics[width=0.9\columnwidth]{final_figs/medical.png}
% \vspace{-10pt}
% \caption{Number of interpretable neurons with Broden categories vs epochs. We see that as training progresses the number of interpretable neurons in higher level categories like 'object' and 'scene' decreases while it increases in lower level categories like 'material' and 'texture'. This shows the transition of the model from being trained on ImageNet to gradually learning to represent the concepts of the new dataset.}
%  \label{fig:medical_stats}
% \end{figure}

\section{Ablation study}
In this section we study the effects of varying the temperature and threshold parameters.
\subsection{Temperature ($T$)}
\begin{figure}[H]
\centering
    \includegraphics[width=0.95\columnwidth]{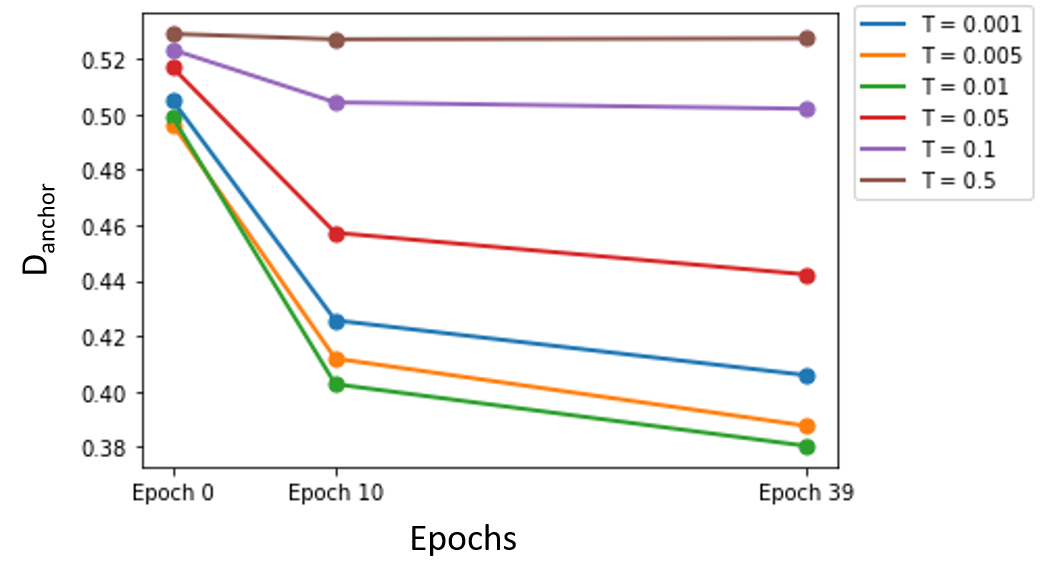}
    \caption{Variation of \textit{anchor distance} across training epochs for different values of temperature in a Resnet18 model trained on Places365. It can be seen that for higher values of temperature such as $T=0.5$ the \textit{anchor distance} is almost constant which shows that the calculated embeddings have lost semantic understanding by weighing all concepts equally. We can see that  variation in anchor loss is maximized for our chosen value $T=0.01$, values lower that $0.01$ show less variation likely caused by the embedding being too focused on the top concept.}
    \label{fig:temperature-training}
\end{figure}
Intuitively from Equation \eqref{eq:embedding}, the temperature parameter decides which concepts to consider for calculating a neuron’s embeddings. Lower temperature implies that the Concept Detector is more confident in its prediction of concepts and just uses the top concepts to calculate the embeddings while a very high temperature implies lower confidence and weighs all words in the concept set equally. To study this we plot the variation of anchor distance for different values of temperatures across training epochs in Fig \ref{fig:temperature-training}. We consider the same Resnet18 model trained on Places365 as Section 3 for consistency of results.

From the figure we see that the anchor metric shows small or no variation for very low or very high temperature values respectively. We think this behavior is expected as for very low temperatures the embedding calculation only considers a single concept while for very high temperatures all concept words in the dataset are weighted equally. Therefore, we choose $T=0.01$ for our experiments as its between the two extremes and captures the variation in semantics across training properly.

We additionally plot the embedding space of the final epoch of the Resnet18 model analyzed in section 3 for different values of temperatures in Fig \ref{fig:temperature-embedding}.

\subsection{Threshold ($\tau$)}

The parameter $\tau$ defines the similarity threshold above which a neuron is considered interpretable. It is used to calculate the number of interpretable neurons in the bar plots in Fig \ref{fig:standard_training_combined}. Varying $\tau$ changes the interpretability threshold, therefore the model considers more or less neurons to be interpretable if we decrease or increase $\tau$ respectively. To study this, we plot the bar plots as in Fig \ref{fig:standard_training_combined} for varying values of $\tau$ in Fig \ref{fig:ablation-tau}. Even though the number of interpretable neurons changes, we noticed that the conclusions we made in section 3.3 still hold. For example, the later layers are more interpretable and represent more complex concepts in all cases. However, if we make $\tau$ too high, the model doesn’t show any neurons to be interpretable. Finally, we would like to point out that $\tau$ is a concept detector dependent parameter and should be changed/tuned accordingly for different concept detectors. The values here are specifically for CLIP-Dissect \citep{clip-dissect}.

% \section{Concept Monitor with different probing dataset}
% As stated in section 3 our method with CLIP-Dissect is able to work with any probing and concept dataset. We provide most of our analysis using Broden dataset as it contains a collection of different concept images and hence is able to provide much better results as compared to a limited dataset. Here we provide an example of that by using CIFAR-100 training images as the probing dataset to analyze a standard training run as done in section 3.  We now use CIFAR-100 training images to monitor neurons 479 and 256. From the embedding space in Fig \ref{fig:cifar100_standard} we can see that neuron 256 converges to the "Field" anchor. We also look at the highly activating images for each neuron in Fig \ref{fig:cifar100_standard} and see that for neuron 479 the most activating images are tree like structures across the sky which are the most similar images to windmills in the CIFAR-100 dataset. The point of this exercise is that concept monitor as all other model dissection methods is dependent on the probing dataset, yet our results are somewhat robust to this choice. Also if we use CLIP-Dissect we are able to use much larger and diverse datasets since we don't require any labelling of images and can simply use any set of images directly.

\section{\algoname{} with a different Concept set}
As stated in Section 3 our method with CLIP-Dissect as concept detector is able to work with any probing and concept dataset. Even though most of our analysis is based on using Broden dataset as $\mathcal{D}_{probe}$, we provide an example of using \algoname{} with CIFAR100 training images in Section 4 to investigate Lottery Ticket Hypothesis. In this section we provide an example of using a different concept set. For a different concept set than broden labels,  we considered the list of top 20000 most common English words \citep{20k} as a concept set and provide our analysis in Figure \ref{fig:20k} From this figure, we can see that the neurons 373, 168, and 114 converge to the same corresponding anchors as in Fig 2 of section 3. This once again highlights the flexibility of our method to track concept evolution using a user preferred set of concepts.

\begin{figure*}
    \centering
    \includegraphics[width=0.95\textwidth]{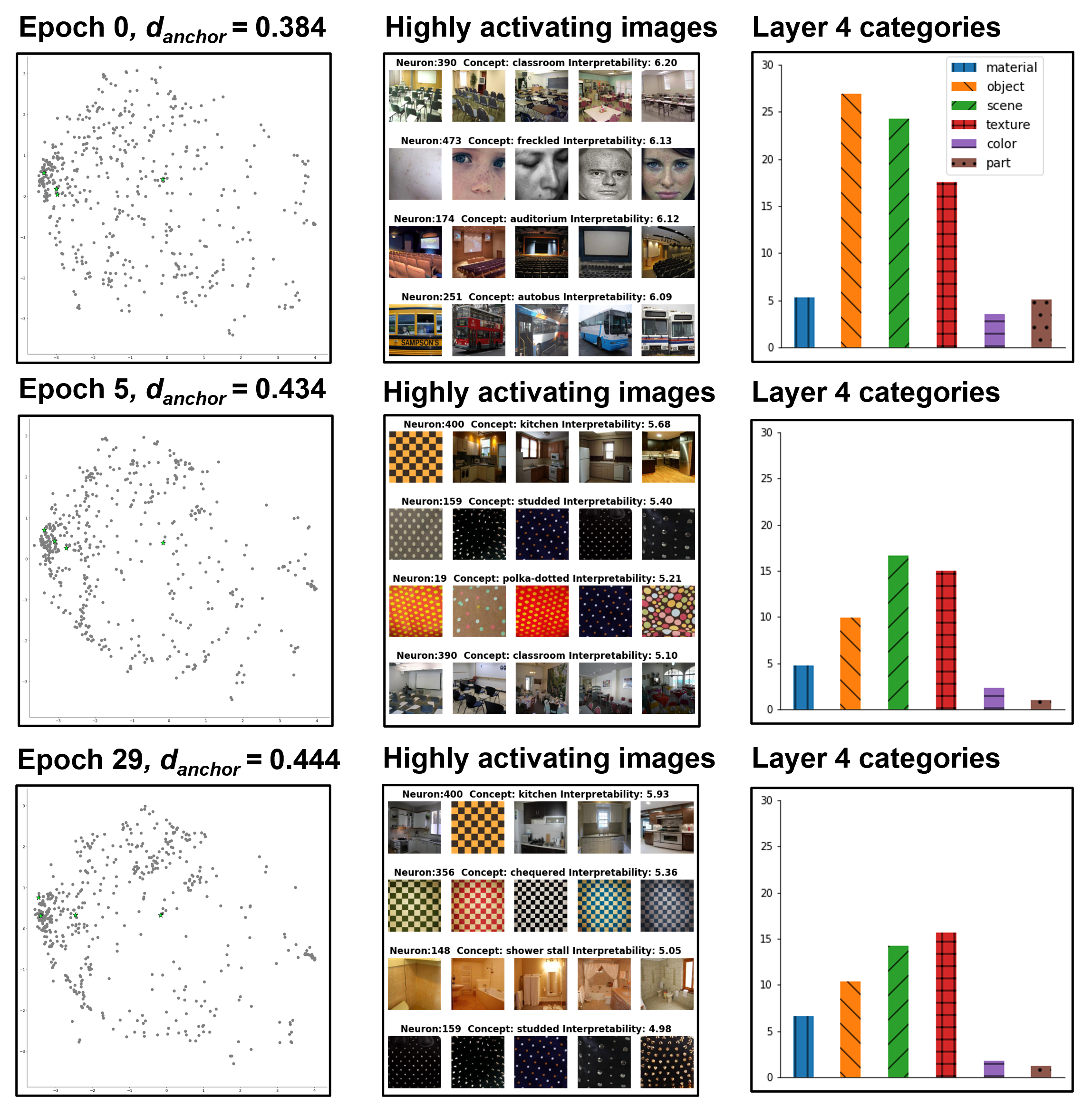}
    \caption{We use \algoname{} to analyze the finetuning a Resnet-34 network on a diabetic retinopathy dataset. On the left we plot our semantic embedding space, where it can be seen that as the training progresses the neuron mass becomes less spread apart, as is also evident from the increasing anchor metric. In the middle we visualize the activating images of \emph{the most interpretable} neurons and we observe that the network starts to focus more on textural aspects. This is also evident from visualizing the category bar plots on the right, where we plot the percentage of interpretable neurons in layer 4 of the network. Each bar represents a different category. We see that the neurons representing complex categories like "object", "scene" decrease while neurons representing "material" and "texture" either remain the same or increase as the training progresses. We also see that texture makes up the majority of interpretable neurons in Epoch 29}
    \label{fig:medical}
\end{figure*}

\begin{figure*}
\centering
    \includegraphics[width=0.9\textwidth]{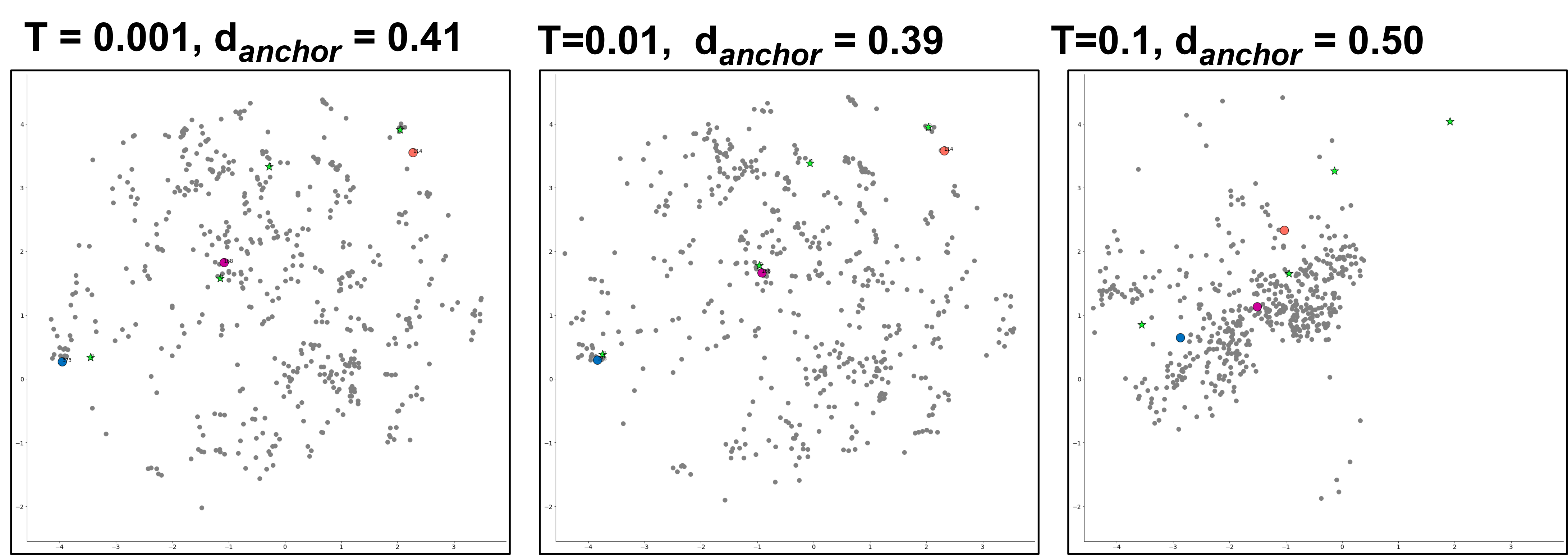}
    \caption{Embedding plots of layer 4 of the final epoch of Resnet18 model plotted for different values of temperature. We see that for very high temperature $T=0.1$, the embedding plot clusters in the same region which confirms our hypothesis that for very high temperatures, the space loses its semantic meaning. This is also represented by $d_{anchor}$. For very low temperature, we see that even though the embedding plot retains its semantic structure, the $d_{anchor}$ is higher, which may be the result of focusing on just the top concept which can provide incomplete understanding of the behavior of a neuron especially in the case of polysemantic neurons.}
    \label{fig:temperature-embedding}
\end{figure*}

\begin{figure*}
\centering
    \includegraphics[width=0.9\textwidth]{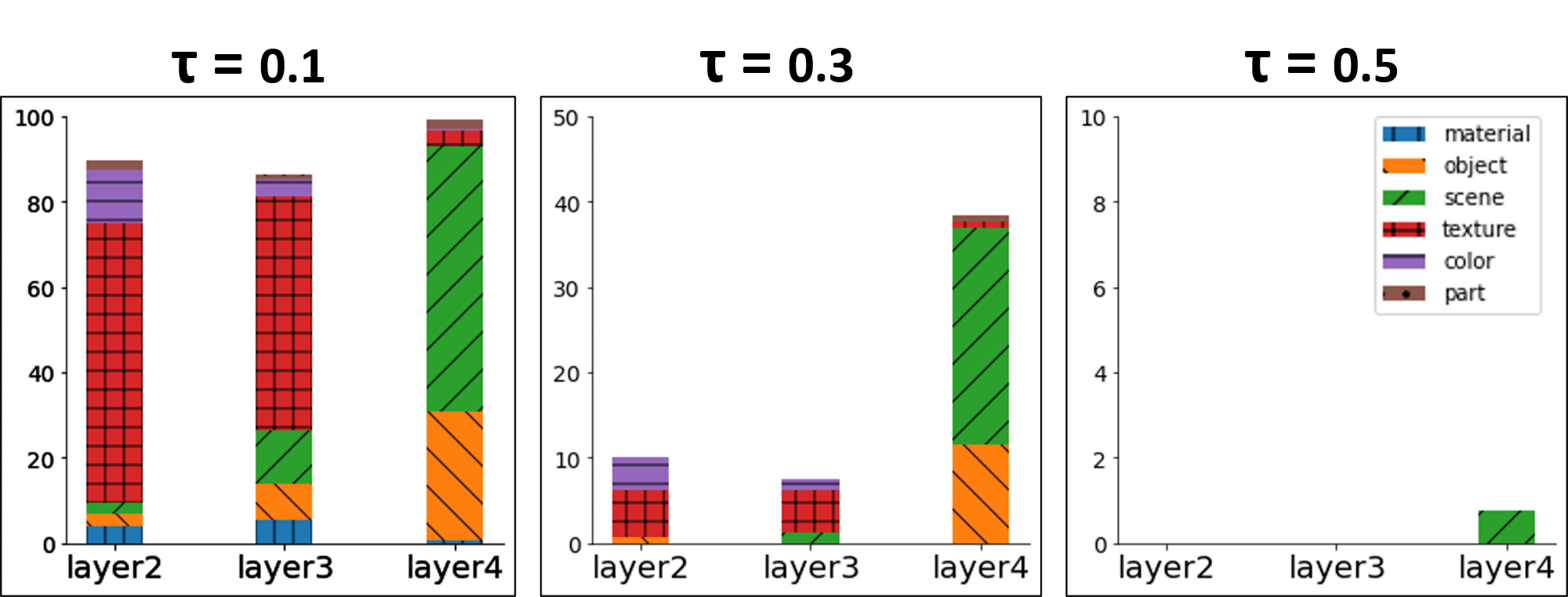}
    \caption{Percentage of interpretable neurons per category for Epoch 39 of a Resnet18 model for different values of  $\tau$. We see that the number of interpretable neurons decreases as we increase the threshold, which is to be expected as we now have a much stricter definition of what counts as "interpretable". We see that even though the number of interpretable neurons decrease as we increase the threshold, we are able to make the same conclusions for most values of $\tau$. For example, for both $\tau=0.1\, 0.3$, we see that later layers represent more complex concepts relating to "object" and "scene" categories as compared to earlier layers which learn "texture" and other simpler features. We also see that if we increase $\tau$ too much we lose all interpretable neurons in some layers. }
    \label{fig:ablation-tau}
\end{figure*}

\begin{figure*}
    \centering
    \includegraphics[width=0.9\textwidth]{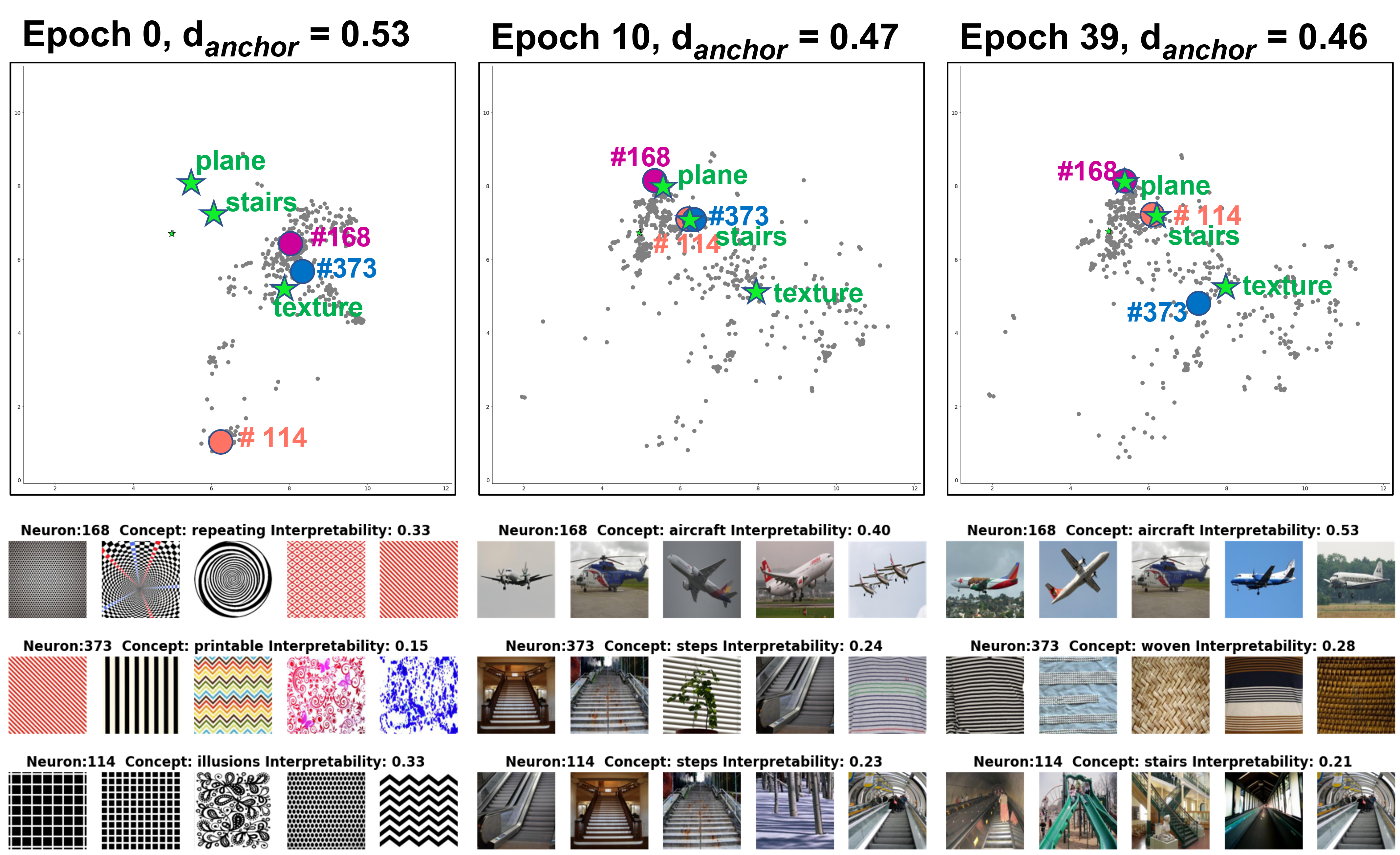}
    \caption{Analysis of Resnet-18 model trained on Places365 dataset using top 20,000 English words \citep{20k} as the concept set. The top image shows our unified embedding space, in which we track Neurons 168, 373, and 114 of layer 4 with the help of semantic anchors "plane", "stairs" and "texture". We see that the trajectory of neurons is exactly the same as what we found with using Broden labels as concept set in section 3. This shows that our method is flexible to using a different concept set and opens the opportunity for users to use their own domain specific concept set to track specific models.}
    \label{fig:20k}
\end{figure*}

\begin{figure*}
\centering
    \includegraphics[width=0.9\textwidth]{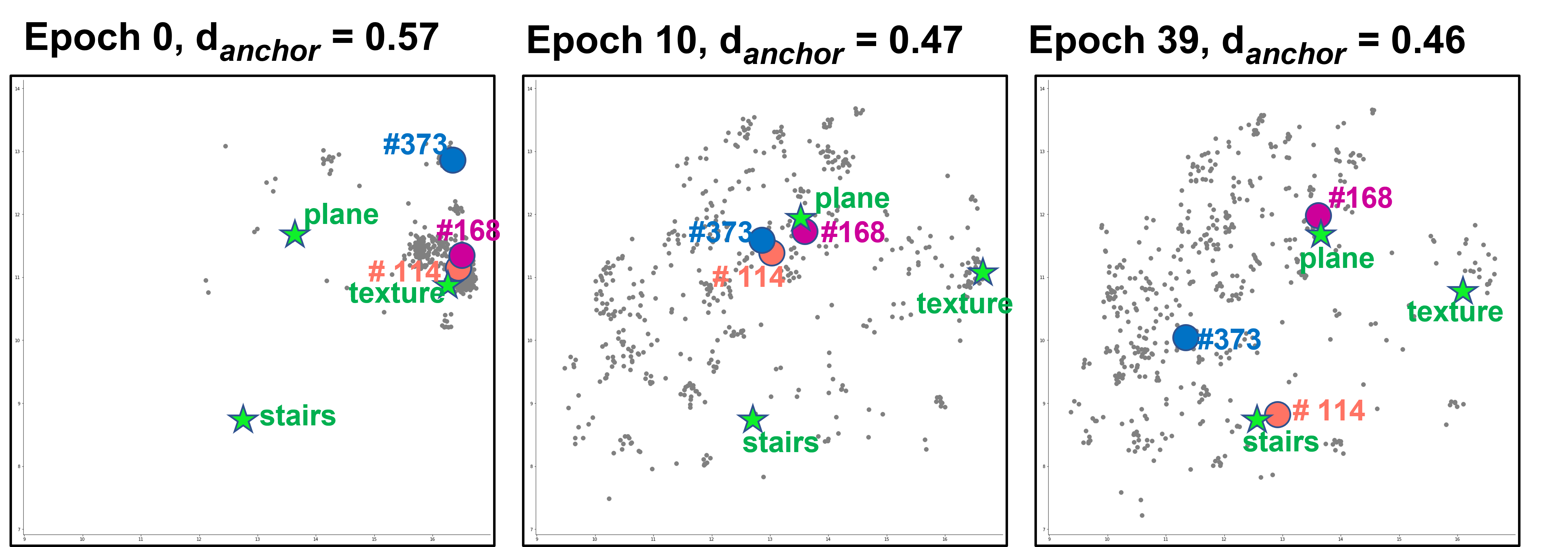}
    \caption{Analysis of Resnet-18 model trained on Places365 using Network Dissection. Here we track neurons 114, 168 and 373 in the layer 4 of the model using our semantic embedding space. We also add concept anchors "plane", "texture" and "stairs" to track the neurons. We see that the neurons start together in a cluster and move towards their learnt concept as the training progresses. We also see that the evolution of neurons is the same as with Clip-Dissect in Section 3 Fig \ref{fig:standard_training_combined} except in the case of neuron 373, which moves away from the "texture" anchor. We attribute this distance to the difference in semantic labelling by Network Dissection, which labels Neuron 373 as "ampitheatre" instead of a textural concept. }
    \label{fig:netdissect}
\end{figure*}

\begin{figure*}
\centering
    \includegraphics[width=0.9\textwidth]{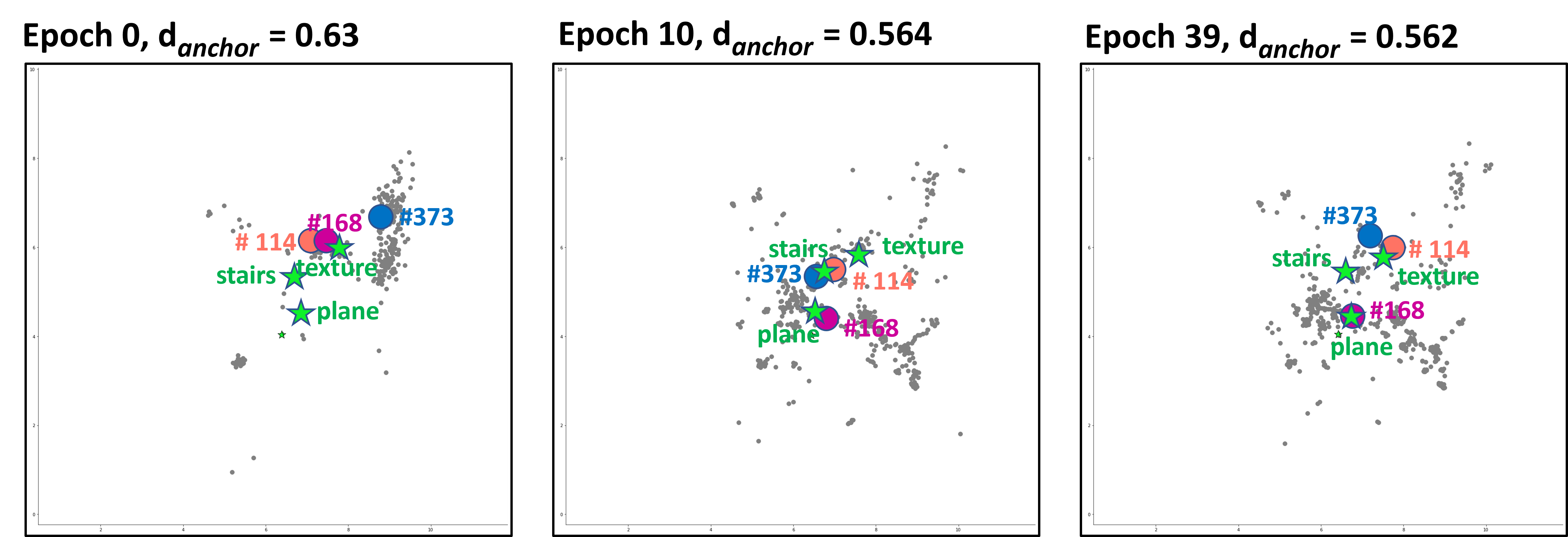}
    \caption{Analysis of Resnet-18 model trained on Places365 using MILAN. Here we track neurons 114, 168 and 373 in the layer 4 of the model using our semantic embedding space. We also add concept anchors "plane", "texture" and "stairs" to track the neurons. We see that the neurons start together in the center and move towards their learnt concept as the training progresses. We also see that the evolution of neurons is exactly the same as with CLIP-Dissect in Section 3 Fig \ref{fig:standard_training_combined} which once again highlights the flexibility of \algoname{} to be used with different concept detectors.}
    \label{fig:milan}
\end{figure*}

\end{document}